\documentclass{llncs}
\usepackage{graphicx}
\usepackage{todonotes}
\usepackage{comment}
\usepackage{enumitem}
\usepackage{microtype}
\usepackage{setspace}
\usepackage[hyphens]{url}
\usepackage{hyperref}
\title{Video Analytics on IoT devices: A Whitepaper}

\author{Sree Premkumar\inst{1} \and
Vimal Premkumar\inst{2} \and
Rakesh Dhakshinamurthy\inst{3}}

\institute{Zoho Corporation, Chennai, India \newline sreeprem1998@gmail.com \and 
Hindustan Institute of Technology and Science\\
\email{vimalprem2003@gmail.com} \and  Dublin City University \\
\email{rakesh.dhakshinamurthy2@mail.dcu.ie}}

\begin{document}
\maketitle
\begin{abstract}
Deep Learning (DL) combined with advanced model optimization methods such as \emph{RC-NN} \cite{rcenniot2020} and \emph{Edge2Train} \cite{edge2trainiot2020} has enabled offline execution of large networks on the IoT devices. In this paper, we compare the modern Deep Learning (DL) based video analytics approaches with the standard Computer Vision (CV) based approaches and finally discuss the best-suited approach for video analytics on IoT devices. 


\end{abstract}

\section{Introduction} \label{Intro}

Today's DL models when efficiently deployed can convert normal IoT devices into intelligent IoT devices that can solve a wide variety of problems. For example, in \cite{aics19smartspeaker}, a face recognition algorithm was trained using Deep Neural Network and deployed on their modern Alexa smart speaker prototype. This model, without disturbing the smart speaker routine, can detect and identify a human face and start the Alexa voice service only when an authorized face is present in the live video frames. Similarly, in \cite{aivision}, a DL and Open CV based object detection model was deployed in their smart speaker. Here, whenever the user calls out the command \emph{Alexa, ask Friday what she sees}, the smart speaker camera turns on and execute the deployed model and calls out the names of detected objects as a response to the user’s command.

As described, since the functionalities of the latest and state-of-the-art devices are based on Machine Learning (ML), the question becomes \emph{Has DL made the traditional CV techniques obsolete?} and \emph{Has ML frameworks superseded traditional CV methods?} This paper will provide a comparison of DL with the more classical handcrafted feature definition approaches in the CV domain. There has been great progress in DL in recent years. Hence, we cannot capture DL sub-domains that solve a variety of real-world problems. In this paper, we shall review traditional algorithmic approaches in CV and brief the applications in which they have been used as a substitute for DL, to complement DL, and to tackle problems that DL cannot handle.

\section{Comparing Deep Learning and Computer Vision}

Rapid advances in DL and IoT hardware capabilities (higher memory,  computing power, image sensor resolution, etc.) have improved the performance and cost-effectiveness of various IoT edge devices. Compared to traditional CV methods, DL enables CV engineers to achieve greater accuracy in tasks such as semantic segmentation, image classification, object detection, etc. Since DL models are trained rather than programmed, applications using the DL models require less expert analysis and fine-tuning. DL also provides better flexibility because models and frameworks can be re-trained using any use case dataset (opposed to highly domain-specific CV methods).

Traditionally, well-established CV techniques such as SIFT \cite{sift}, SURF \cite{surf}, BRIEF \cite{brief}, etc. (feature descriptors) are used in object detection IoT use cases. Before the emergence of DL models and frameworks, feature extraction was used as the first step for image classification tasks, where features are descriptive or informative patches in images. Multiple CV algorithms, such as corner detection, edge detection, threshold segmentation, etc., are involved in this first step. These features extracted from images form a definition of each object class. At the deployment stage, other images are searched for these definitions. If a significant number of known features are found in the target image, the target image is identified as a chair, horse, etc. The challenge with this CV approach is that it is necessary to choose the important in each given image. As the number of image classes increases, feature extraction becomes more and more cumbersome. It is up to the skill of a CV engineer’s judgment and a long trial and error process to find the best features that describe different classes of objects.

With all the top approaches in CV, the workflow of the CV engineer has dramatically changed. i,e., \emph{the knowledge and expertise in extracting hand-crafted features have been replaced by knowledge and expertise in iterating through deep learning architectures}. We depict this in Fig. 1. The development of ML had tremendously influenced the field of CV and is responsible for a big jump in the ability to detect objects. This progress has been enabled by an increase in computing power, and the amount of data available for training ML models. The recent explosion in and wide-spread adoption of various deep-neural network architectures for CV is apparent from the high citation of DL papers such as ImageNet Classification with Deep Convolutional Neural Networks.
`
\begin{figure}
\centering
 \includegraphics[width=12cm,height=7cm,keepaspectratio]{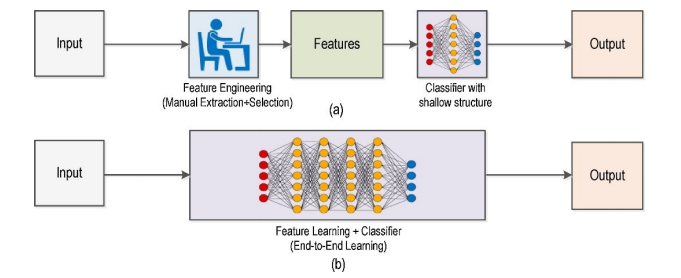}
\caption{(a) Traditional Computer Vision workflow vs. (b) Deep Learning workflow \cite{o2019deep}.}
\label{dl_vs_cv}  
\end{figure}

\section{Computer Vision Advantages for IoT Devices}

Here we describe the feature descriptors that can improve the CV-based video analytics tasks on IoT devices.

SIFT and SURF feature descriptors need to be generally combined with traditional machine learning classification algorithms such as SVMs \cite{covidaway2020}, SVRs \cite{wfiot2020}, KNearest Neighbours to solve the CV problems.  DL is sometimes overkilling when the given problem is simple and if it can be solved by CV techniques. Algorithms like pixel counting, SIFT, simple color thresholding are not class-specific. i.e., they are very general and perform the same for any number of images. In opposition, features learned from a DL model are specific to a training dataset. If the dataset is not well constructed, the model would not perform well for unseen images. Therefore, we recommend using SIFT and other algorithms for basic applications such as image-stitching, which does not need specific class knowledge. One needs to use experience, and also common sense when selecting the route to take for a target CV application. For example, to classify two classes of products in a factory floor conveyor belt, an ML model will work given that enough data can be collected for training. However, the same classification can be achieved when simple color thresholding is realized. By finding such alternatives, a resource-friendly solution can be developed and deployed on IoT devices.

\section{Challenges of Deep Learning on IoT Devices}

There are multiple challenges when implementing DL approaches on IoT devices. Although the DL models show high performance, there is a high computation and power cost that is required by billions of additional math operations. DL models require high resources for training and sometimes also for inference. Dedicated hardware such as high-powered GPUs and TPUs are mandatory for training large models, and AI accelerated platforms such as VPUs for inference, which highly increases the cost of IoT devices. Achieving satisfactory performance in object classification requires high-resolution images or video data, and more importantly, DL-based vision processing tasks also depend on image resolution. This image or video frame resolution is particularly important for IoT applications where we need to detect and classify objects in the long distance. 

In such resource-demanding applications, frame reduction techniques such as SIFT features or optical flow for moving objects need to be used as the first step to identify a region of interest, then perform the required object detection task. DL needs big training data with millions of data rows with proper labeling. For example, Microsoft Common Objects in Context (COCO) consists of 2.5 million images with 91 object categories, ImageNet consists of 1.5 million images with 1000 object classes, and PASCAL VOC Dataset consists of 500K images with 20 object classes.      

\section{Efficient Video Analytics on IoT Devices}

The models optimized using methods such as \emph{Edge2train} and \emph{RCE-NN} can run on the IoT devices. Processing data at the edge level without depending on the cloud improves latency, reduces subscription \& cloud storage costs, processing requirements, and bandwidth requirements. It also can address privacy and security issues by avoiding the transmission of sensitive or identifiable data over the network. 

Hybrid or composite approaches involving conventional CV and DL should be used to take great advantage of such heterogeneous computing abilities available at the edge. Such heterogeneous systems consist of a combination of multiple processors and chipsets. For example, in the Smart Hearing Aid prototype from \cite{smarthearingaid}, the users have integrated a DSP based microphone array with a Linux SBC to perform edge level audio processing such as noise suppression, the direction of arrival estimation, etc., without depending on the internet. The IoT devices can be power efficient when the user assigns different workloads to the most efficient compute engine. During the design phase, users have to design a hybrid approach that is a combination of DL with hand-crafted feature extractors. For example, when designing a facial-expression recognizing IoT application, a new feature loss can be used \cite{zeng2018hand}. Here, the information of hand-crafted features is embedded into the network training process in order to reduce the difference between hand-crafted features and features learned by the DL model. 

At the edge level, simple problems such as automatic panorama stitching, video stabilization, 3D modeling, motion estimation, scene understanding, motion capture, video processing should be solved using basic CV techniques and not by training DL models. DL should be supplemented by other CV techniques that will make us reach artificial general intelligence and we need to identify problems where DL shows poor performance and should be supplemented with the classical CV techniques.

\section{Discussion}

Old CV techniques have become obsolete in recent years because of DL. In this paper, we have laid down many arguments for why traditional CV techniques are still very much useful even in the age of DL. We have compared CV and DL and discussed how sometimes traditional CV can be considered to be a more resource-friendly alternative in situations where DL is overkill for a specific task. Although DL models are driving exciting breakthroughs in the domain of IoT analytics, we reviewed how traditional CV techniques can improve DL performance in a wide range of applications.

\begin{spacing}{0.92}
\bibliography{aics-sample.bib}
\bibliographystyle{splncs03}
\end{spacing}

\end{document}